\documentclass[letterpaper, 10 pt, conference]{ieeeconf}  % Comment this line out if you need a4paper

\usepackage{booktabs}

\IEEEoverridecommandlockouts                              % This command is only needed if 
\usepackage{graphicx}
\usepackage{url}  %IVM added this because of the URLs in the reference list
\usepackage{pifont}
\usepackage{comment}
\usepackage{amsmath}

\usepackage{lipsum}
\usepackage{amssymb}

\title{\LARGE \bf
SEMT: Static-Expansion-Mesh Transformer Network Architecture\\ for Remote Sensing Image Captioning }

\author{Khang~Truong,
        Lam~Pham,
        Hieu~Tang,
        Jasmin~Lampert,
        Martin~Boyer,
        Son~Phan,
        Truong~Nguyen
\thanks{L. Pham, M. Boyer,  and J.Lampert are with Austrian Institute of Technology, Vienna, Austria}
\thanks{H.Tang is with University of technology of Troyes, France}
\thanks{K. Truong and T. Nguyen are with Ho Chi Minh University of Technology, Vietnam} 
\thanks{Son Phan is with Ton Duc Thang University, Vietnam} 
}

\begin{document}

\maketitle
\thispagestyle{empty}
\pagestyle{empty}

%%%%%%%%%%%%%%%%%%%%%%%%%%%%%%%%%%%%%%%%%%%%%%%%%%%%%%%%%%%%%%%%%%%%
\begin{abstract}
Image captioning has emerged as a crucial task in the intersection of computer vision and natural language processing, enabling automated generation of descriptive text from visual content. 
In the context of remote sensing, image captioning plays a significant role in interpreting vast and complex satellite imagery, aiding applications such as environmental monitoring, disaster assessment, and urban planning. 
This motivates us, in this paper, to present a transformer based network architecture for remote sensing image captioning (RSIC) in which multiple techniques of Static Expansion, Memory-Augmented Self-Attention, Mesh Transformer are evaluated and integrated.
We evaluate our proposed models using two benchmark remote sensing image datasets of UCM-Caption and NWPU-Caption.
Our best model  outperforms the state-of-the-art systems on most of evaluation metrics, which demonstrates potential to apply for real-life remote sensing image systems.

%This paper presents an overview of recent advancements in image captioning techniques tailored for remote sensing data. 
%We explore the challenges posed by high-resolution imagery, domain-specific terminology, and the need for fine-grained descriptions. 
%Furthermore, we evaluate state-of-the-art deep learning models, including convolutional neural networks (CNNs) and transformer-based architectures, highlighting their effectiveness in generating accurate and contextually rich captions. 
%Experimental results demonstrate the potential of these approaches to enhance situational awareness and decision-making processes in remote sensing applications. 
%Finally, we discuss future research directions, emphasizing the importance of multi-modal data fusion, domain adaptation, and the integration of geographical knowledge into captioning systems.

\indent \textit{Items}--- vision language, remote sensing image, image captioning, transformer.
\end{abstract}
%%%%%%%%%%%%%%%%%%%%%%%%%%%%%%%%%%%%%%%%%%%%%%%%%%%%%%%%%%%%%%%%%%%%%%%%%%%%%%%%
\section{INTRODUCTION}
The rapid growth of remote sensing technologies has resulted in the accumulation of vast amounts of satellite imagery, offering opportunities for Earth observation and analysis. 
However, the effective interpretation and extracting meaningful insights from remote sensing images remain a challenging task. 
Remote sensing image captioning, which aims to automatically generate descriptive textual information from images, has emerged as a promising solution to bridge the gap between visual data and human understanding.
In recent years, significant progress has been made in the field of remote sensing image captioning, particularly through the development of deep learning models with transformer-based architectures. 
Indeed, various deep learning models such as \cite{mlca}, \cite{rs_capnet}, \cite{rs-vit}, \cite{clip-rsicd}, \cite{glcm}, \cite{sat}, \cite{smatt}, \cite{struc-att}, \cite{mc-net}, \cite{vrtmm}, \cite{p-to-h} have been proposed and demonstrated remarkable success in the task of remote sensing image captioning (RSIC).
These network architectures can be separated into two main groups.
The first group~\cite{rs_capnet, rs-vit, clip-rsicd} utilized pre-trained models that were trained on large-scale image datasets. 
These models are then finetuned on target datasets of remote sensing images.
In other words, this approach leverages transfer learning techniques to reduce training time and the cost of network construction, but still achieves the potential and competitive models.
However, this approach results in high model complexity, as it leverages part or all of entire large pre-trained models.

Instead of leveraging the pre-trained models, the second group~\cite{glcm, sat, smatt, struc-att, mc-net, p-to-h} focuses on exploring advances of network architecture that shows potential to improve the model performance.
For examples, the proposed model in~\cite{glcm} explored a CNN-based network in the encoder component to capture local and global features of remote sensing images.
Similarly, a CNN-based architecture was proposed in~\cite{mc-net} to capture multi-scale contextual features.
Meanwhile,~\cite{struc-att, smatt, p-to-h} made efforts to construct innovative attention layers to effectively capture distinct features from remote sensing images.

Although these recently published models proposed for the RSIC task have achieved promising performance, these models mainly were constructed basing on encoder-decoder architectures, especially traditional transformer-based networks.
This motivates us to evaluate recently innovative transformer architectures that is potential to further improve the RSIC task performance.
In other words, we are inspired by the second approach, focusing on exploring innovative and advanced network architecture to construct an effective model for RSIC task.
In particular, three techniques of Mesh Transformer~\cite{meshed_memory} (Mesh Trans.),  Memory-Augmented Attention (Mem. Att.)~\cite{meshed_memory} and Static Expansion (Stat. Exp.)~\cite{static_expansion} are evaluated and integrated to construct our proposed models.
The proposed models are evaluated on two benchmark datasets of UCM-Caption~\cite{ucm_data} and NWPU-Caption~\cite{mlca}, and then compared with the state-of-the-art models.

\section{Proposed Transformer Based Network Architecture}

% Proposed model:  meshed memory transformer:  backbone (b), encoder (e), and decoder (d).
\begin{figure*}[t]
    \centering
    \includegraphics[width=1.0\linewidth]{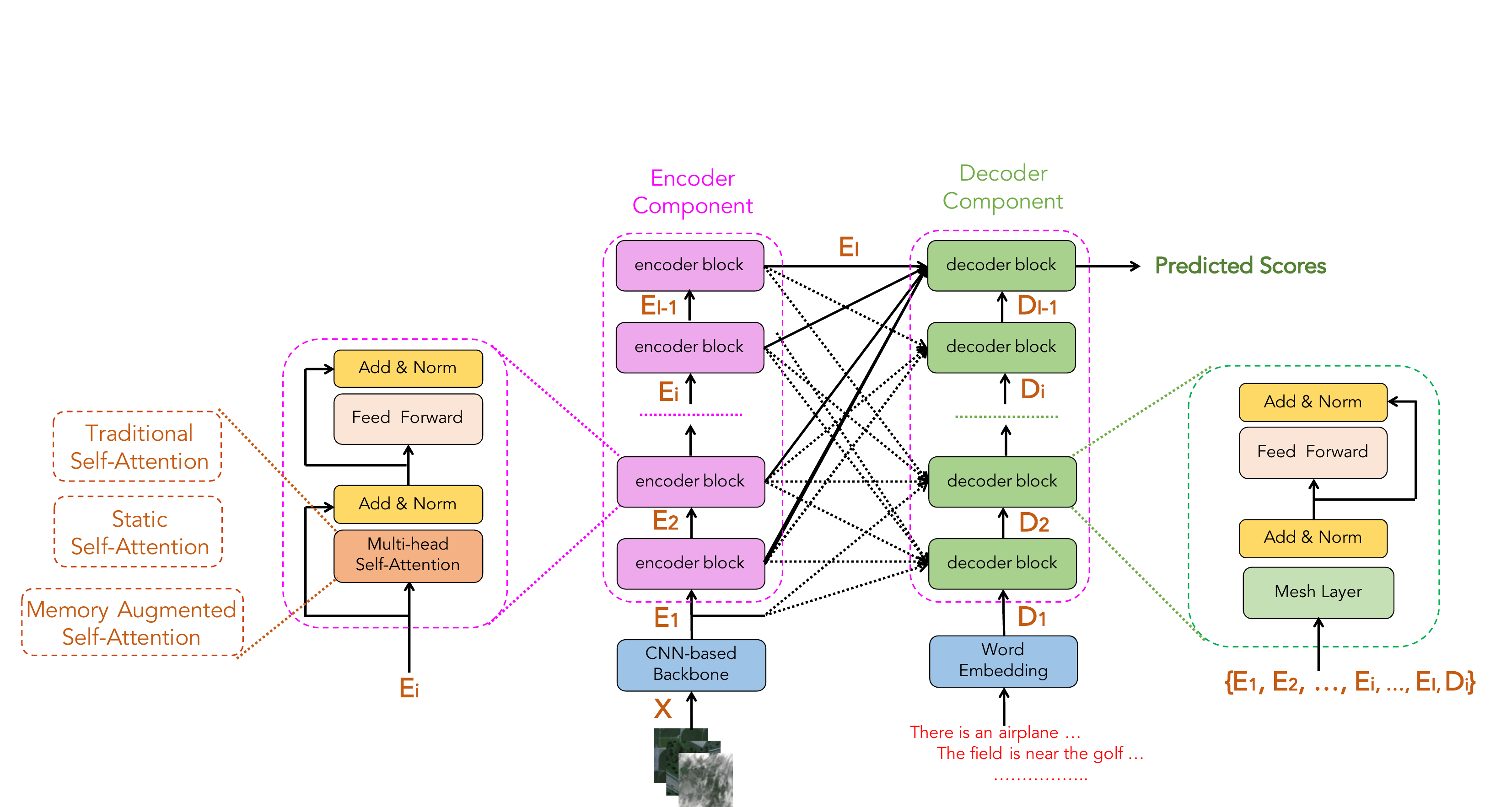}
    \caption{The high-level architecture of the proposed models with mesh-based transformer architecture}
    \label{fig:high-level model}
\end{figure*}

The high-level architecture of the proposed networks is shown in the Fig.~\ref{fig:high-level model} that includes 4 main components: CNN-based Backbone, Word Embedding, Encoder, and Decoder. 
In particular, the input images $\mathbf{X}$ are first fed into the CNN-based Backbone to extract image feature maps $\mathbf{E_1}$.
Then, the image feature maps go through the Encoder component, generating $\{\mathbf{E_2, E_3, .... E_I}\}$ as the outputs of encoder blocks in the Encoder component.
Meanwhile, the input captions are fed into the Word Embedding to extract word embeddings $\mathbf{D_1}$.
The word embeddings $\mathbf{D_1}$ then go through the Decoder component.
As the proposed model is based on the Mesh Transformer architecture~\cite{meshed_memory}, all outputs from encoder blocks in the Encoder component and the output of CNN-based Backbone, $\{\mathbf{E_1, E_2, .... E_I}\}$, are sent to all decoder blocks in the Decoder component.
The output of each decoder block in Decoder component is denoted by $\mathbf{D_i}$.
\begin{table}[t]
 \caption{Hyper-parameters set for the proposed models}
  \centering
  \begin{tabular}{lc}
    \toprule
    Parameters & Values \\
    \midrule
    Training epoch number & 20 \\
    Learning rate  & $10^{-4}$ \\
    Learning rate decay ratio (per epoch) & 0.95 \\
    Batch size & 500 \\
    The longest length of captions ($L$) & 53 \\
    Word embedding dimension ($E$) & 768 \\
    Number of encoder blocks & 4 \\
    Number of decoder blocks & 4 \\
        \bottomrule
  \end{tabular}
  \label{tab:setting}
\end{table}
\subsection{Word Embedding}
As shown in Fig.~\ref{fig:high-level model}, the Word Embedding component receives the captions (e.g., each caption is an array of discrete words), each of caption presents the fixed length $L$ which is the length of the longest caption in the evaluating dataset.
Then, the captions are fed into the positional encoding, generating word feature maps each of which is denoted by $\mathbf{D_1} \in \mathbb{R}^{L{\times}E}$, where $E$ is the dimension of embedding vectors representing for words.
In other words, the positional encoding is used to transform discrete words into numerical vectors with $E$ dimension.
Finally, the word feature maps are fed into the Decoder component.

\subsection{CNN-based Backbone}
In this paper we evaluate various CNN-based Backbone of VGG16, MobileNet-V2, Resnet152, Inception, and EfficientNetB2 architectures. 
To construct these CNN-based Backbone, these pre-trained models of VGG16, MobileNet-V2, Resnet152, Inception, and EfficientNetB2, which were trained on ImageNet dataset~\cite{imagenet_ds}, are leveraged.
Then, dense layers of these pre-trained models are removed to obtain the CNN-base Backbones.

The CNN-based backbones are used to transform an input image $\mathbf{X} \in \mathbb{R}^{H{\times}W{\times}C}$ into an image feature map of $\mathbf{X_f} \in \mathbb{R}^{H_f{\times}W_f{\times}C_f}$, where $H, W, C$ are height, width, channel of the original image and $H_f, W_f, C_f$ are height, width, channel of the image feature map.
Then, the $H_f$ and $W_f$ dimensions are flatten to generate a feature map of $\mathbf{E_1} \in \mathbb{R}^{F_f{\times}C_f}$ before feeding into the Encoder component, where $F_f = W_f.H_f$.

%They are used to extract features from images.  The images that is fed to the model have shape $256\times256\times3$.  When batched, the tensor has shape $N \times 256 \times 256 \times 3$.  The feature map has shape $N \times 8 \times 8 \times D$ where $D$ is the dimension of the backbone output (Resnet $D=2048$, VGG16 $D = 512$, Inception $D = 760$, Efficient Net $D = 1408$, Mobile Net$(D = 1280)$.  Flatten: Nx64xD --> encoder

% Output of encoder (5 output as using mesh architecture, each output has the same shape Nx64x768

%After that the tensor is flattened to the shape $N \times S \times D$ for feeding to the multihead attention where $S = 64$ and $D = 768$ for this paper.  
\begin{table*}[t]
 \caption{Performance comparison between our proposed models and the state-of-the-art systems on NWPU-Caption dataset}
  \centering
  
  \begin{tabular}{lcccccc}
    \toprule
    Models     & BLEU-1 & BLEU-2 & BLEU-3 & BLEU-4 & METEOR & ROUGE-L \\
    \midrule

    GLCM~\cite{glcm} & 0.554 & 0.423 & 0.335 & 0.272 & 0.279 & 0.504 \\
    SAT \cite{sat} & 	0.734 &	0.612 & 	0.528 & 0.469 & 0.337& 0.601 \\
    SM-Att \cite{smatt} & 0.739 & 0.616 & 0.534 & 0.469 & 0.338	& 0.595 \\
    Struc-Att \cite{struc-att} & 0.744 & 0.609 &	0.519 &	0.456 & 0.309 &	0.606 \\
    MC-Net \cite{mc-net} & 0.741 & 0.626 & 0.544 & 0.478 & 0.347 & 0.611 \\
    MLCA-NET \cite{mlca} & 0.745 & 0.624 & 0.541 & 0.478 & 0.337 & 0.601\\
    P-to-H \cite{p-to-h} & 0.757 & 0.629 & 0.546 & 0.483 & 0.319 & 0.586 \\
    RS-ViT-B \cite{rs-vit}& 0.810 & - & - & 0.547 & - & -\\
    VRTMM \cite{vrtmm} & 0.811 & 0.703 & 0.621 &	0.557 & 0.366 & 0.685 \\
    CLIP-RSICD \cite{clip-rsicd} & 0.826 & - & -& 0.565 & - & - \\
    CLIP-Cap-4 \cite{rs_capnet} & 0.871 & - & - & 0.650 & - & - \\
    RS-CapRet \cite{rs_capnet} & 0.871	&0.786	& 0.713	& 0.650 & 0.439& 0.775 \\
    RS-CapRet-finetuned \cite{rs_capnet} & 0.871 & 0.787&\textbf{0.717}&\textbf{0.656}&0.436& 0.776 \\
    
    \hline
    Proposed M1 (Trad. Att.) & 0.870 & 0.772 & 0.686 & 0.608 & 0.447 & 0.782 \\
    Proposed M2 (Trad. Att., Mesh Trans.)  & 0.869 & 0.772 & 0.686 & 0.608 & 0.445 & 0.782 \\
    Proposed M3 (Trad. Att., Mesh Trans., Mem. Att.)   & 0.854 & 0.755 & 0.671 & 0.594 & 0.440 & 0.770\\
    Proposed M4 (Stat. Att.)             & 0.872 & 0.772 & 0.687 & 0.609 & 0.440 & 0.777 \\    
    Proposed M5 (Stat. Att., Mesh Trans.)  - \textbf{SEMT}             & \textbf{0.882} & \textbf{0.792} & 0.711 & 0.636 & \textbf{0.453} & \textbf{0.793} \\
    
   \bottomrule
  \end{tabular}
  \label{tab:res_03}
\end{table*}

\begin{table*}
\centering
\caption{Performance comparison between our proposed models and the state-of-the-art systems on UCM-Caption dataset} % Main caption
\label{tab:performance_metrics_partial} % You can use any label you prefer
\begin{tabular}{lcccccc} % Adjusted for 1 method column + 6 data columns
\hline
Model & BLEU 1 & BLEU 2 & BLEU 3 & BLEU 4 & METEOR & ROUGE-L \\
\hline
SAA \cite{ref114} & 0.783 & 0.728 & 0.676 & 0.633 & 0.380 & 0.686 \\
TrTr-CMR \cite{ref97} & 0.816 & 0.709 & 0.622 & 0.547 & 0.398 & 0.744 \\
STA \cite{ref113} & 0.817 & 0.779 & 0.613 & 0.606 &  - &  - \\
GLCM \cite{ref111} & 0.818 & 0.754 & 0.699 & 0.647 & 0.462 & 0.752 \\
CHA \cite{ref87} & 0.823 & 0.768 & 0.710 & 0.659 &  - & 0.756 \\
Deformable \cite{ref96} & 0.823 & 0.770 & 0.723 & 0.679 & 0.444 & 0.784 \\
MLCA-Net \cite{ref83} & 0.826 & 0.770 & 0.717 & 0.668 & 0.435 & 0.772 \\
GVFGA \cite{ref110} & 0.832 & 0.766 & 0.710 & 0.660 & 0.444 & 0.785 \\
Adaptive \cite{ref93} & 0.839 & 0.769 & 0.715 & 0.675 & 0.446 &  - \\
MSMI \cite{ref89} & 0.843 & 0.775 & 0.711 & 0.651 & 0.453 & 0.785 \\
MC-Net \cite{ref84} & 0.845 & 0.784 & 0.732 & 0.679 & 0.449 & 0.786 \\
SCAMET \cite{ref95} & 0.846 & 0.777 & 0.726 & 0.681 & 0.526 & 0.817 \\
MAN \cite{ref92} & 0.848 & & & 0.664 & 0.456 & 0.807 \\
RASG \cite{ref109} & 0.852 & 0.793 & 0.743 & 0.698 & 0.457 & 0.807 \\
\hline
Proposed M1 (Trad. Att.) & 0.839 & 0.789& 0.744& 	0.701& 0.506 & 0.826\\
Proposed M2 (Trad. Att. Mesh Trans) & 0.845 & 0.787& 0.737 &0.688& 0.490& 0.810 \\
Proposed M3 (Trad. Att., Mesh Trans, Mem. Att.) & 0.864& 0.810& \textbf{0.764}& \textbf{0.717} & \textbf{0.499} & \textbf{0.827} \\
Proposed M4 (Stat. Att.) & 0.868 & 0.809 & 0.755 & 0.703 & 0.491 & 0.821 \\
Proposed M5 (Stat. Att., Mesh Trans.)  - \textbf{SEMT}  & \textbf{0.871} & \textbf{0.812} & 0.760 & 0.710 & 0.493 & 0.826 \\
\bottomrule
\end{tabular}
  \label{tab:res_03_02}
\end{table*}

\subsection{Encoder Component}
As shown in Fig.~\ref{fig:high-level model}, the Encoder component is constructed by multiple encoder blocks, each of which comprises one multihead self-attention layer and one feedforward layer together with normalization layers.
Regarding the techniques used to construct the multihead self-attention layer in this paper, we not only evaluate Traditional Self-Attention~\cite{trad_att} (Trad. Att.), but we also evaluate two recently innovative attention techniques: Memory-Augmented Self-Attention (Mem. Att.) inspired by~\cite{meshed_memory} and Statistic Expansion (Stat. Exp.) inspired from~\cite{static_expansion}, respectively.

Given the input $\mathbf{E_i} \in \mathbb{R}^{2}$, the output $\mathbf{O} \in \mathbb{R}^{2}$ of the Traditional Self-Attention (Trad. Att.) is computed by:
\begin{equation}
\begin{matrix}
    \mathbf{O} = Attention(\mathbf{Q}, \mathbf{K}, \mathbf{V}) \\ 
    = softmax(\frac{\mathbf{Q}.\mathbf{K^T}}{\sqrt{d_k}}).\mathbf{V}, where \\
    \mathbf{Q} = \mathbf{W_q}.\mathbf{E_i}, \\
    \mathbf{K} = \mathbf{W_k}.\mathbf{E_i}, \\
    \mathbf{V} = \mathbf{W_v}.\mathbf{E_i},
\end{matrix}
\end{equation}
and $\mathbf{W_q}$, $\mathbf{W_k}$, $\mathbf{W_v} \in \mathbb{R}^{2}$ are matrices of learnable weights; $d_k$ is a constant to define one dimension of $\mathbf{Q}$ and $\mathbf{K}$.

Regarding the Memory-Augmented Self-Attention (Mem. Att.), this technique reuses the formula of the Traditional Self-Attention (Trad. Att.), but $\mathbf{K}$ and $\mathbf{V}$ in the Traditional Self-Attention are added by a learnable matrix.
Adding a learnable matrix into $\mathbf{K}$ and $\mathbf{V}$ aims to overcome the limitation of the Traditional Self-Attention which cannot model a priori knowledge on relationships between image regions.
In other words, a learnable matrix is considered as an additional “slots”  where a priori information is encoded.
By this way, relationships between image regions are learn effectively.
As a result, the output $\mathbf{O}\in\mathbb{R}^{2}$ of the Memory-Augmented Self-Attention (Mem. Att.) layer is performed by:
\begin{equation}
\begin{matrix}
    \mathbf{O} = Attention(\mathbf{Q}, \mathbf{K}, \mathbf{V}), where \\
    \mathbf{Q} = \mathbf{W_q}.\mathbf{E_i}, \\
    \mathbf{K} = [\mathbf{W_k}.\mathbf{E_i}, \mathbf{M_k}], \\
    \mathbf{V} = [\mathbf{W_v}.\mathbf{E_i}, \mathbf{M_v}],
\end{matrix}
\end{equation}
and $\mathbf{M_k}$, $\mathbf{M_v}\in\mathbb{R}^{2}$ are are learnable matrices used to add into $\mathbf{K}$ and $\mathbf{V}$.

%Query $Q \in \mathbb{R}^{N\times64\times D}$. This query is multiplied with a learnable parameter $\mathcal{P} \in \mathbb{R}^{L_s \times D}$ where $L_s=300$ is an arbitrary number represent the size of the memory. This forms $M \in R^{N\times L_s \times 64}$, a type of attention-score that represents the compatibility with the learnt memory. The score $M$ is multiplied with the Query, generating $\mathcal{E} \in R^{N\times L_s \times D}$, the result sequence of the forward pass. To perform the backward pass, which is to produce a tensor that has a similar shape of the input, by multiply $\mathcal{E}$ with $M^T$ it creates the output $\mathcal{O}$, which has the shape $N\times64 \times 768$.

Regarding the Static Expansion, this technique considers the input feature map as a sequence and processes the sequence in two phases.
In the first phase, referred to as the forward operation, the input sequence is distributed using an increased or arbitrary number of elements which are the learnable parameters.
Then, the input sequence is retrieved to the original length in the second phase, referred to as the backward operation.
By using the forward-backward mechanism, this enforces the network to learn the relevant and sequential features in the input sequence more effectively.
Additionally, the output of Static Expansion layer also has the same shape with the input sequence that fits the attention mechanism in a transformer architecture.
Given the input $\mathbf{E_i} \in \mathbb{R}^{2}$ as the input matrix of the Static Expansion layer, the output $\mathbf{O} \in \mathbb{R}^{2}$ of the Static Expansion layer is computed by:
\begin{equation}
\begin{matrix}
    \mathbf{M} = Normalize_{L2}(Relu(\mathcal{\mathbf{E_i}}.\mathbf{P})), \\
 \mathbf{I} = \mathbf{M}.\mathbf{E_i^T}, \\ 
 \mathbf{O} = \mathbf{M^T}.\mathbf{I^T}, \\
\end{matrix}
\end{equation}
where $\mathbf{P} \in \mathbb{R}^{2}$ is the matrix of trainable parameters that is used to expand the sequence length of the input $\mathbf{E_i}$. 

%\subsubsection{Traditional transformer mechanism}
%The encoder takes advantage of the transformer encoder structure, with the exception that all the features extracted from the encoder layers are used, instead of the final one as it is in transformer architecture.  Given a set of image regions $X^0$ extracted from an input image, attention can be used to obtain a permutation invariant encoding of $X^0$ through the self-attention operations used in the Transformer. Noticeably, attentive weights depend solely on the pairwise similarities between linear projections of the input set  itself. Therefore, the self-attention operator can be seen as a way of encoding pairwise relationships inside the input set. By this way, the encoder learns how to interpret different features of the feature map, without maintaining the absolute position.  In this scenario, the first layer receives a flattened feature map with shape $N\times 64 \times D$, it then performs self-attention, then return the output. As described above, the output of layer $i$ is $X^i$. All the output are captured and return, transfer to the decoder. Therefore, this decoder layer return a list of 5 tensor, $\mathcal{X} = \{X^0, X^1, ...X^4\}$.

\begin{table}[t]
 \caption{\textbf{Ablation}: Evaluate the CNN-based Backbone architectures NWPU-Caption dataset}
  \centering
  \scalebox{0.7}{

  \begin{tabular}{lllllll}
    \toprule
    %& \multicolumn{6}{c}{SEMT Model} \\
    \cmidrule(r){2 - 7}
    Model (backbone)     & BLEU 1 & BLEU 2 & BLEU 3 & BLEU 4 & METEOR & ROUGE-L \\
    \midrule   
    SEMT (ResNet152)  & 0.868 & 0.774 & 0.691 & 0.615 & 0.444 & 0.780\\
    SEMT (VGG16)  & 0.871 & 0.768 & 0.679 & 0.598 & 0.428 & 0.766 \\
    SEMT (Inception)  & 0.850& 0.741 & 0.651 & 0.570 & 0.424 & 0.752 \\
    SEMT (EfficientNetB2)  & \textbf{0.882} & \textbf{0.792} & \textbf{0.711} & \textbf{0.636} & \textbf{0.453} & \textbf{0.793} \\
    SEMT (MobileNet-V2)  & 0.861 & 0.760 & 0.672 & 0.593 & 0.436 & 0.769 \\
    \bottomrule
  \end{tabular}}
  \label{tab:res_01}
\end{table}
\begin{table}[t]
 \caption{\textbf{Ablation}: Evaluate the number of multihead attention on NWPU-Caption dataset}
  \centering
    \scalebox{0.7}{
  \begin{tabular}{lllllll}
    \toprule
    & \multicolumn{6}{c}{EfficientNetB2 Backbone} \\
    \cmidrule(r){2 - 7}
    Models     & BLEU 1 & BLEU 2 & BLEU 3 & BLEU 4 & METEOR & ROUGE-L \\
    \midrule   
    SEMT (4 heads)  & 0.872 & 0.780 & 0.699 & 0.624 & 0.449 & 0.787 \\
    SEMT (8 heads)  & \textbf{0.882} & \textbf{0.792} & \textbf{0.711} & \textbf{0.636} & \textbf{0.453} & \textbf{0.793} \\
    SEMT (12 heads)  & 0.880 & 0.787 & 0.707 & 0.636 & 0.459 & 0.793 \\
    SEMT (16 heads)  & 0.875 & 0.781& 0.698 & 0.622 & 0.448 & 0.783 \\
    SEMT (20 heads) & 0.866 & 0.772& 0.689& 0.614& 	0.450 &	0.785\\
    SEMT (24 heads)  & 0.871 & 0.780 & 0.699& 0.626 & 0.458 & 0.791\\
    \bottomrule
  \end{tabular}}
  \label{tab:res_02}
\end{table}

\begin{table}[t]
\caption{\textbf{Ablation}: Evaluate the CNN-based Backbone architectures 
 on UCM-Caption dataset}
\centering
\scalebox{0.7}{
 \begin{tabular}{lllllll}
  \toprule
  %& \multicolumn{6}{c}{SEMT Model} \\
     %\cmidrule(r){2 - 7}
    Model (backbone)    & BLEU 1 & BLEU 2 & BLEU 3 & BLEU 4 & METEOR & ROUGE-L \\
    \midrule
    SEMT (Resnet152) & 0.846 & 0.778 & 0.723 & 0.671 & 0.469 & 0.803\\
    SEMT (VGG16) & 0.841 & 0.781 & 0.730 & 0.682 & 0.476 & 0.799\\
    SEMT (Inception) & 0.801 & 0.740 & 0.692 & 0.647 & 0.465 & 0.767\\
    %SEMT (Short CNN) & 0.619 & 0.506 & 0.429 & 0.360 & 0.311 & 0.576\\
    SEMT (EfficientNetB2) & \textbf{0.871} & \textbf{0.812} & \textbf{0.760} & \textbf{0.710} & \textbf{0.493} & \textbf{0.826} \\
    SEMT (MobileNet-V2) & 0.851 & 0.791 & 0.740 & 0.692 & 0.487 & 0.811\\
    \bottomrule
    \end{tabular}}
      \label{tab:res_01_02}
\end{table}

\begin{table}[t]
\caption{\textbf{Ablation}: Evaluate the number of multihead attention on UCM-Caption dataset}
 \centering
 \scalebox{0.7}{
 \begin{tabular}{lllllll}
 \toprule
 & \multicolumn{6}{c}{EfficientNetB2 Backbone} \\
 \cmidrule(r){2 - 7}
 Models& BLEU 1 & BLEU 2 & BLEU 3 & BLEU 4 & METEOR & ROUGE-L \\
 \midrule
SEMT (4 heads) & 0.865 & 0.807 & 0.757 & 0.710 & 0.492 & 0.823 \\
 SEMT (8 heads) & \textbf{0.857} & \textbf{0.799} & \textbf{0.750} & \textbf{0.704} & \textbf{0.498} & \textbf{0.821} \\
 SEMT (12 heads) & 0.853 & 0.797 & 0.752 & 0.709 & 0.506 & 0.824 \\
 SEMT (16 heads) & 0.856 & 0.796 & 0.747 & 0.700 & 0.494 & 0.819 \\
 SEMT (20 heads) & 0.871 & 0.812 & 0.760 & 0.710 & 0.493 & 0.826 \\
 SEMT (32 heads) & 0.848 & 0.790 & 0.741 & 0.696 & 0.493 & 0.813 \\
\bottomrule
 \end{tabular}}
   \label{tab:res_02_02}
\end{table}

\begin{comment}
The encoder take the flattened feature map ($N\times S \times D$) from the backbone as the first input, namely $X^0$. We accumulate through each encoder layer and get $X^1, X^2, ..., X^n$. The purpose of the encoder is to extract all relevant information from all encoder layers, then pipe to the decoder layer. The encoder output format should have the form $N\times (NumEncoder + 1) \times S \times D$. \\
Formally, as proposed in \cite{meshed_memory}, the meshed memory encoder output has the following format:
$$\mathcal{X}=\left(X^n, X^{n-1},..., X^0\right)$$
where 
$$X_{i+1}=\mathcal{M}_{mem}\left(X^i\right)$$
$$\mathcal{M}_{mem}\left(X\right)=Attn\left(XW_q, concat[X,M_K],concat[X,M_V]\right)$$
In this context, $X^0$ is the input embedding, taken from the CNN backbone, $M_K,M_V$ denote the learnable parameters. This makes $\mathcal{X}$ the set of all encoder layers outputs.
\end{comment}

\subsection{Decoder Component}
% Input 1: 5 shapes from encoder
% Input 2: caption : array of words - token (token number basing on dataset) - embedding (position emb)

% Output: array of vector, each vector is one-hot format, each vector size: 3000 x 53/70 --> check code???

% Decoder: 5 layers that equals to the number of output encoder.

% + Input, output, functions (f1, f2, ..., fn)
% + f1 la 1 network --> hinh ve
% + Trong f1 co mesh function (m)
% cong thuc cua m

% Copied from Meshed memory
As shown in Fig.~\ref{fig:high-level model}, Decoder component comprises multiple decode blocks, each of which comprises one Mesh layer and one feed-forward layer.
It can be seen that the difference between the proposed Decoder component and that of a traditional transformer lies on the Mesh layer where all encoder blocks' outputs and CNN-Backbone output are merged as using the Mesh Transformer architecture.
The inspiration of the mesh-based transformer is to enforce the Decoder component to exploit both low-level and high-level feature maps generated by the Encoder component.
This avoids the lost of information at higher decoder blocks in the Decoder component when many decoder blocks are configured in the network.

Given $\{\mathbf{E_1, E_2, ..., E_i,..., E_I}\}$ as the outputs from the CNN-based Backbone ($\mathbf{E1}$) and from $I-1$ encoder blocks in the Encoder component (from $\mathbf{E_2}$ to $\mathbf{E_I}$) at each decoder block $i$, and the input sequence from the previous decoder block is denoted $\mathbf{D}_{i-1} \in \mathbb{R}^{2}$, the operation in the Mesh layer comprises four main steps in the order: (1) Self-Attention applied for input sequence $\mathbf{D}_{i-1}$, (2) Cross-Attention over Multi-Level Feature Maps, (3) Gating with Sigmoid Fusion, and finally (4) Aggregation.
These steps are described below:
%\mathbb{R}^{2N \times 5 \times L \times D}$, where $N$ is the batch size, 5 is the number of tokens per sequence (or spatial locations), $L$ is the sequence length, and $D$ is the hidden dimension.
\subsubsection{Self-Attention applied for input sequence}

We first apply a multi-head self-attention mechanism to the input sequence $\mathbf{D}_{i-1}$
\begin{equation}
\begin{matrix}
    \mathbf{D_a} = Attention(\mathbf{Q}, \mathbf{K}, \mathbf{V}), where \\
    \mathbf{Q} = \mathbf{W_q}.\mathbf{D_{i-1}}, \\
    \mathbf{K} = \mathbf{W_k}.\mathbf{D_{i-1}}, \\
    \mathbf{V} = \mathbf{W_v}.\mathbf{D_{i-1}},
\end{matrix}
\end{equation}
and $\mathbf{W_q}$, $\mathbf{W_k}$, $\mathbf{W_v} \in \mathbb{R}^{2}$ are matrices of learnable weights, $\mathbf{D_a}$ is the output of the attention layer.
%where $\text{Attention}_\text{self}$ is parameterized by its own set of learnable weights for query, key, and value projections. This operation produces the refined sequence $\mathbf{S}_a \in \mathbb{R}^{N \times 5 \times L \times D}$.

\subsubsection{Cross-Attention over Multi-Level Features}

Next, the cross-attention between the input sequence $\mathbf{D_a}$ with the outputs of Encoder component and CNN-based Backbone $\{\mathbf{E_1, E_2, ..., E_i,..., E_I}\}$ is computed by:

\begin{equation}
\begin{matrix}
    \mathbf{T_i} = Attention(\mathbf{Q}, \mathbf{K}, \mathbf{V}), where \\
    \mathbf{Q} = \mathbf{W_q}.\mathbf{D_{a}}, \\
    \mathbf{K} = \mathbf{W_k}.\mathbf{E_{i}}, \\
    \mathbf{V} = \mathbf{W_v}.\mathbf{E_{i}}, \\
    \quad \text{for } i = 1, \ldots, I,
\end{matrix}
\end{equation}
and $\mathbf{W_q}$, $\mathbf{W_k}$, $\mathbf{W_v} \in \mathbb{R}^{2}$ are matrices of learnable weights,  $\mathbf{T_i}$ is the output of the self-attention layer.
%$$
%\mathbf{T}_k = \text{Attention}_\text{cross}(\mathbf{S}_a, \mathbf{E}_k, \mathbf{E}_k), \quad \text{for } k = 1, \ldots, I,
%$$
%where $\text{Attention}_\text{cross}$ uses a separate set of learnable parameters, shared across all encoder levels $\mathbf{E}_k$.

\subsubsection{Gating with Sigmoid Fusion}

For each $\mathbf{T}_i$, a gating mechanism is applied to learn its contribution by computing a weight map via a pointwise gating function:

\begin{equation}
\begin{matrix}
   \mathbf{R}_i = Sigmoid\left(\mathbf{W_\text{gate}} \cdot \textit{Concat}(\mathbf{T}_i, \mathbf{D}_a) + \mathbf{b_\text{gate}} \right),
\end{matrix}
\end{equation}
%%\mathbf{R}_k = \sigma\left(W_\text{gate} \cdot \text{concat}(\mathbf{T}_k, \mathbf{S}_a) + b_\text{gate} \right),
where $\mathbf{W_\text{gate}} \in \mathbb{R}^{2}$ and $\mathbf{b_\text{gate}} \in \mathbb{R}$ are learnable projection weights. 
%The concatenated input has shape $N \times 5 \times L \times 2D$, and the output $\mathbf{R}_k \in \mathbb{R}^{N \times 5 \times L \times D}$ serves as an attention gate for each token.

\subsubsection{Aggregation}
Finally, the gated encoder contributions are aggregated through element-wise multiplication and summation:
\begin{equation}
\begin{matrix}
    \mathbf{O} = \sum_{i=1}^{I} \mathbf{R}_i \odot \mathbf{T}_i
\end{matrix},
\end{equation}
%\mathbf{S} = \sum_{k=1}^{I} \mathbf{R}_k \odot \mathbf{T}_k,
where $\odot$ denotes pointwise product, and $\mathbf{O}$ is the output of the Mesh layer.

%\subsubsection{Final Feedforward Projection}

%The final output of the decoder stage is obtained through a two-layer feedforward network with ReLU activation:

%$$
%\mathbf{D}_i = W_{f_2} \cdot \text{ReLU}(W_{f_1} \cdot \mathbf{S} + b_1) + b_2,
%$$
%where $W_{f_1}, W_{f_2} \in \mathbb{R}^{D \times D}$, and $b_1, b_2 \in \mathbb{R}^{D}$ are learnable parameters.

\begin{figure}[t]
    \centering
    \includegraphics[width=1.0\linewidth]{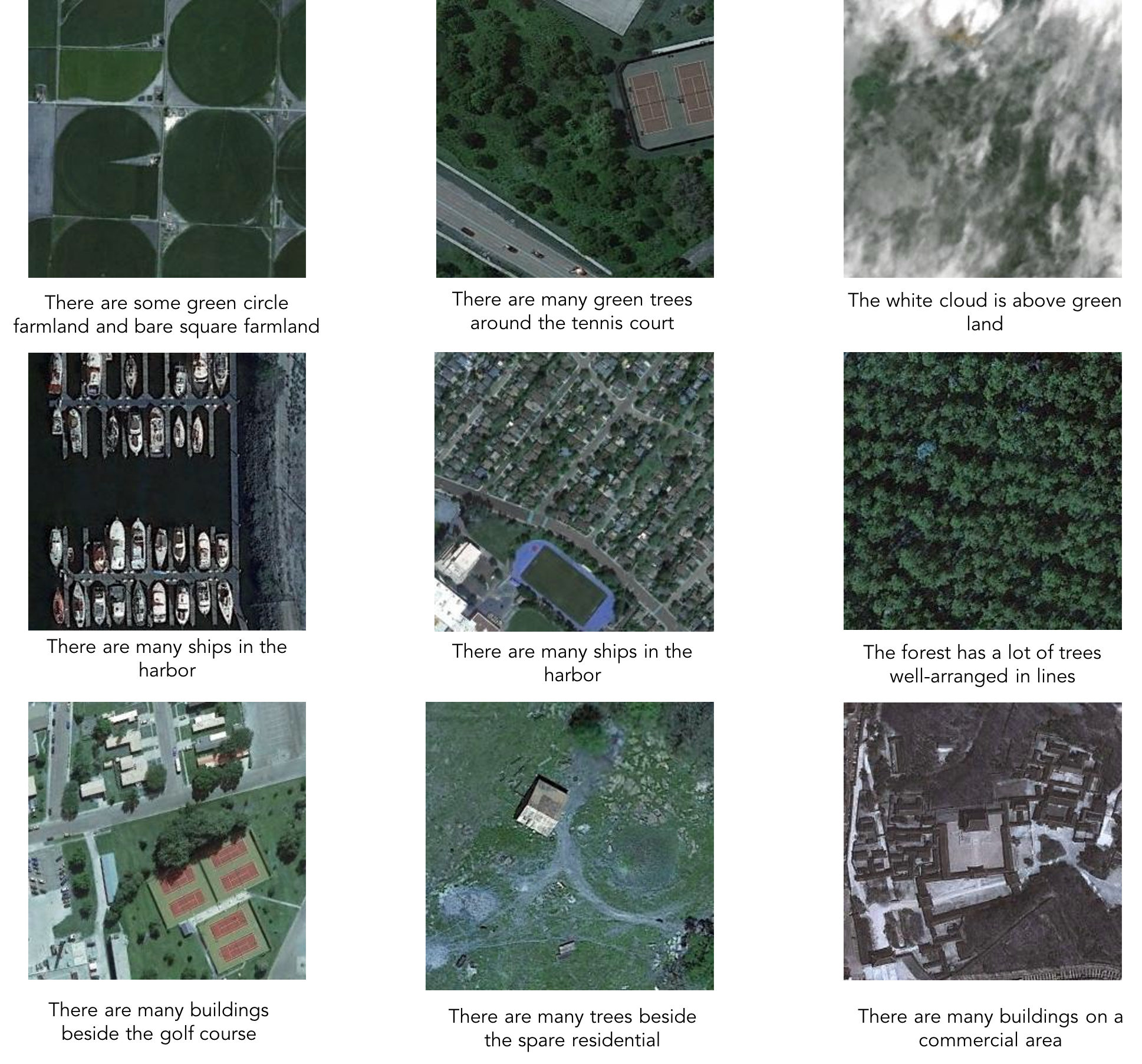}
    \caption{Caption results for some remote sensing images from Test subset in NWPU-Caption dataset using the proposed SEMT system}
    \label{fig:res_cap}
\end{figure}

\section{Experimental Results and Discussion}
\subsection{Dataset}
In this paper, the proposed models are evaluated on two benchmarch datasets of UCM-Caption~\cite{ucm_data} and NWPU-Caption~\cite{mlca}. 

UCM-Caption~\cite{ucm_data} is the earliest publicly dataset which was proposed for the task of remote sensing image captioning.
UCM-Caption dataset contains 2100 images which are from 21 categories.
Each image presents the size of $256\times256$ and together with 5 different caption sentences.

%which presents a large-scale benchmark dataset specifically designed for remote sensing image captioning. 
The NWPU-Caption~\cite{mlca}, which was recently published in 2022 and currently presents the largest dataset proposed for RSIC task (i.e. A comparison between NWPU-Caption and other datasets proposed for RSIC is presented in~\cite{mlca}).
The NWPU-Caption dataset contains a substantial number of remote sensing images (e.g., 31,500 images with RGB format). 
Each of image is annotated with five natural language sentences, resulting in a total of 157,500 captions. 
The dataset has a rich and varied vocabulary, containing 3,149 unique words collected from the captions. 
The images have a uniform size of $256\times256$. 
The spatial resolution of the images varies, ranging approximately from 0.2 meter to 30 meters.

We follow the data splitting in the papers~\cite{ucm_data, mlca} that proposed UCM-caption and NWPU-Caption datasets respectively and present the results on Test set (10\% of the entire dataset).

\subsection{Evaluation Metrics}
We evaluate the proposed models in this paper using metrics of BLEU-1, BLEU-2, BLEU-3, BLEU-4, METEOR, and ROUGE-L.
These metrics, considered as standard metrics in generative tasks, have been widely used in the remote sensing image captioning~\cite{mlca, rs_capnet, glcm}.

\subsection{Experimental settings}
Our proposed models were constructed with Tensorflow framework and trained on TPUv2-8.
 The first 5 epochs are trained at constant learning rate. After that, the rate decays exponentially. 
We used Adam~\cite{adam} method for the optimization. 
Regarding the hyper-parameters set in the proposed models, Table ~\ref{tab:setting} presents the initial learning rate, the learning rate decay ration, the number encoder and decoder, respectively.

\subsection{Experimental results and discussion}
To evaluate multiple techniques of Static Expansion (Stat. Exp.), Memory-Augmented Self-Attention (Mem. Att.), Mesh Transformer (Mesh Trans.), we constructed five different transformer-base network configurations: (1) M1 model with Traditional Self-Attention and without using any innovation technique; (2) M2 model with Traditional Self-Attention and Mesh Transformer technique; (3) M3 model with Traditional Self-Attention, Memory-Augmented Self-Attention, and Mesh Transformer; (4) M4 model with only Static Expansion; and finally (5) M5 model with Static Expansion and Mesh Transformer techniques.
All these models use EfficientNetB2 as the CNN-based Backbone and 8-head setting at the multi-head attention layers in both Encoder and Decoder components.

 Experimental results on NWPU-Caption dataset shown in the lower part of Table~\ref{tab:res_03} indicate that applying Static Expansion technique (e.g. M4 model) is more effective to further improve RSIC performance compared with other models using Traditional Self-Attention (e.g., M1, M2, and M3 models).
When we combine both Static Expansion and Mesh Transformer techniques in M5 model, we achieve the best model configuration.
We referred M5 model to as SEMT and compare this model with the state-of-the-art systems.
As the results are shown in the upper part in Table~\ref{tab:res_03}, our best model, SEMT, outperforms the state-of-the-art systems on most of evaluation metrics (BLEU-1, BLEU-2, METEOR, ROUGE-L) and is competitive on the metrics of BLEU-3 (top-3), BLEU-4 (top-4).

Regarding the results on UCM-Caption dataset as shown in Table~\ref{tab:res_03_02}, our proposed models (M3 and M5) outperform the state-of-the-art systems on all evaluation metrics. The high performance on both benchmark datasets prove our proposed systems robust for the remote sensing image captioning task.

We also conduct experimental experiments to evaluate roles of CNN-based Backbone and the number of heads in multi-head attention layers.
In particular, while the Word Embedding, Encoder, and Decoder architectures are remained from experiments in Table~\ref{tab:res_03}, we evaluate different CNN-based Backbone of VGG16, MobileNet-V3, ResNet152, Inception-V3, and EfficientNetB2. 
The results shown in Table~\ref{tab:res_01} and Table~\ref{tab:res_01_02} indicate that EfficientNetB2 outperforms the other CNN-based Backbone on all metrics.
Given the best performance on EfficientNetB2, we remain this backbone, then evaluate the number for heads used in attention layers in both Encoder and Decoder.
As the results are shown in Table~\ref{tab:res_02} and Table~\ref{tab:res_02_02}, we achieve the best performance with the setting of 8 heads.

As the results shown and discussed, we achieve the best model, SEMT, which combines the techniques of Mesh Transformer \& Static Expansion and use EfficientNetB2 Backbone with 8-head setting for muti-head attention layers.
The Fig.~\ref{fig:res_cap} presents the captioning results from the proposed SEMT model for some remote sensing images in Test subset of NWPU-Caption dataset.
It can be seen that all captions effectively reflect the context of the remote sensing images.

\section{Conclusion}
We have presents a transformer based network architecture for remote sensing image captioning.
By combining both Static Expansion and Mesh Transformer techniques, we successfully constructed the SEMT model, achieving high performance on two benchmark datasets of UCM-Caption and NWPU-Caption.
Our proposed SEMT model outperforms the state-of-the-art models on most of metrics, demonstrating its potential for real-world applications in remote sensing images analysis.

%that proves potential to apply on real-life systems for analyzing remote sensing images.

%\newpage
%Demo
%\begin{figure}[t]
%    	\vspace{-0.2cm}
%    \centering
%    \includegraphics[width =1.0\linewidth]{f2.png}
%	\caption{A demo of landslide segmentation on Huggingface}
%   	%\vspace{-0.4cm}
%    \label{fig:res_f2}
%\end{figure}

%\addtolength{\textheight}{-1cm}   % This command serves to balance the column lengths

%\begin{thebibliography}{99}
\bibliographystyle{IEEEbib}
\bibliography{refs}
%\end{thebibliography}
\end{document}